\documentclass{sig-alternate-05-2015}

\begin{document}

% Copyright
% \setcopyright{acmcopyright}

%\setcopyright{acmlicensed}
%\setcopyright{rightsretained}
%\setcopyright{usgov}
%\setcopyright{usgovmixed}
%\setcopyright{cagov}
%\setcopyright{cagovmixed}

% DOI
%\doi{10.475/123_4}

% ISBN
%\isbn{123-4567-24-567/08/06}

%Conference
\conferenceinfo{Workshop on Surprise, Opposition, and Obstruction in Adaptive and Personalized Systems (SOAP)}{June 13--17, 2016, Halifax, NS, Canada}

%\acmPrice{\$15.00}

%
% --- Author Metadata here ---
%\conferenceinfo{SOAP Workshop}{'16 Halifax, NS, Canada}
%\CopyrightYear{2007} % Allows default copyright year (20XX) to be over-ridden - IF NEED BE.
%\crdata{0-12345-67-8/90/01}  % Allows default copyright data (0-89791-88-6/97/05) to be over-ridden - IF NEED BE.
% --- End of Author Metadata ---

\title{Towards Playlist Generation Algorithms Using\\RNNs Trained on Within-Track Transitions\titlenote{
%(Produces the permission block, and
%copyright information). For use with
%SIG-ALTERNATE.CLS.
% Supported by ACM. 
This work was part funded by the ``Fusing Audio and Semantic Technologies for Intelligent Music Production and Consumption'' (FAST IMPACt) EPSRC Grant EP/L019981/1 and the European Commission H2020 research and innovation grant AudioCommons (688382). Sandler acknowledges the support of the Royal Society as a recipient of a Wolfson Research Merit Award.
}}

%\subtitle{[Extended Abstract]
%\titlenote{A full version of this paper is available as \textit{Author's Guide to Preparing ACM SIG Proceedings Using \LaTeX$2_\epsilon$\ and BibTeX} at \texttt{www.acm.org/eaddress.htm}}}


\numberofauthors{3} %  in this sample file, there are a *total*
\author{
% 1st. author
\alignauthor
Keunwoo Choi \\
       \affaddr{Centre for Digital Music}\\
       \affaddr{EECS, QMUL}\\
       \affaddr{London, UK}\\
       \email{keunwoo.choi@qmul.ac.uk}
% 2nd. author
\alignauthor
Gy\"orgy Fazekas \\
       \affaddr{Centre for Digital Music}\\
       \affaddr{EECS, QMUL}\\
       \affaddr{London, UK}\\
       \email{g.fazekas@qmul.ac.uk}% 3rd. author
\alignauthor 
Mark Sandler \\
       \affaddr{Centre for Digital Music}\\
       \affaddr{EECS, QMUL}\\
       \affaddr{London, UK}\\
       \email{m.sandler@qmul.ac.uk}
}

\maketitle
\begin{abstract}
We introduce a novel playlist generation algorithm 
that focuses on the quality of transitions using 
a recurrent neural network (RNN). The proposed model 
assumes that optimal transitions between tracks can be 
modelled and predicted by internal transitions within music tracks.
We introduce modelling sequences of high-level music descriptors 
using RNNs and discuss an experiment involving different similarity 
functions, where the sequences are provided by 
a musical structural analysis algorithm. Qualitative observations show that the proposed approach 
can effectively model transitions of music tracks in playlists.
\end{abstract}



\pretolerance=10000
% \raggedright


\section{Introduction}
\label{intro}

In music recommendation, the quality of transitions become important particularly when the recommendation is provided in the form of a playlist. This is due to a unique aspect of music consumption. Unlike other products, music is consumed \textit{i) instantaneously}, for instance, while listening using streaming services, \textit{ii) repeatedly}, i.e., listeners are willing to listen to the same music multiple times, and \textit{iii) quickly}, i.e., an item usually only lasts a few minutes. Hence, recommended items are typically consumed or played in a sequence. This behaviour introduces the need for \textit{good transitions} between items, that is, the relevance and subjective judgement a recommended track depends on the previous track.

Recommendation approaches using collaborative filtering are prone to overlook niche or new items, although the popularity bias of known items can be compensated for. This is called the \textit{cold-start problem} \cite{ricci2011introduction}.
Content-based approaches which are designed to solve the cold-start problem can suffer from
lack of diversity when recommended items are selected simply by similarity. This is often called top-$N$ recommendation.
It is well known that \textit{unexpectedness}, \textit{surprise} or \textit{serendipity} play an important role in the music
recommendation and discovery \cite{choi2015understanding}. Compared to other strategies, focusing on
transitions can naturally provide these qualities. 

There have been approaches that primarily focus on the transitions of tracks \cite{liebman2015dj}, \cite{chen2012playlist}, \cite{mcfee2011natural}.
They assumed the \textit{Markov} property of hidden states or embeddings of tracks. Using the Markov property, it is assumed that future events only depend on the current one and does not depend on the past. This has been successfully used for sequence modelling for instance in speech \cite{rabiner1989tutorial} too.
 In music computing, playlist datasets \cite{mcfee2012hypergraph}, \cite{maillet2009steerable}, \cite{chen2012playlist} collaboratively created for reference by DJs and listeners were used for training and evaluation of sequence modelling approaches. Although these datasets consist of a large number of tracks, e.g. 101k playlists in \cite{mcfee2012hypergraph}, the lack of audio data fundamentally limits research based on audio content analysis. 

% NEW
Recently, recurrent neural networks (RNNs) have become widely 
used for sequence modelling in tasks such as speech recognition, 
substantially outperforming previous hidden Markov model-based approaches \cite{sak2014long}. 
The success of the application of RNNs largely relies on the introduction of Long Short-Term Memory (LSTM) units  \cite{gers2000learning}. The merit of LSTM comes from the gate cells of LSTM units, that decide how much the units take input, release output, and forget the previous events. Especially, the forget gate improves the training efficiency by helping the gradients flow well.
However, RNNs have not been used for playlists generation and modelling,
due in part to the lack of sufficient training data. 
To solve this problem, we propose using an RNN trained on \textit{within-track} 
transitions to model playlists. 

We assume that transitions between structural segments of music can be used
as a model for generating the desired high-quality transitions between tracks. In general, 
segments in a track are different but coherent and their musical features can be expected to match well in succession. 
This is due to the careful and intentional design by the composer. 
Using this approach, the number of transitions can easily outnumber that of existing playlist datasets, and therefore it enables to train an RNN model.

The rest of the paper is organised as follows. The proposed method is first described in Section \ref{sec:prop}.
We then present experimental results and discussion in Section \ref{sec:exp} and conclude in Section \ref{sec:conclusinon}.

%===============
% 2. MODEL
%===============

\section{The Proposed Model}
\label{sec:prop}

\begin{figure}[t!]
\begin{center}
\centerline{\includegraphics[width=1.0\columnwidth]{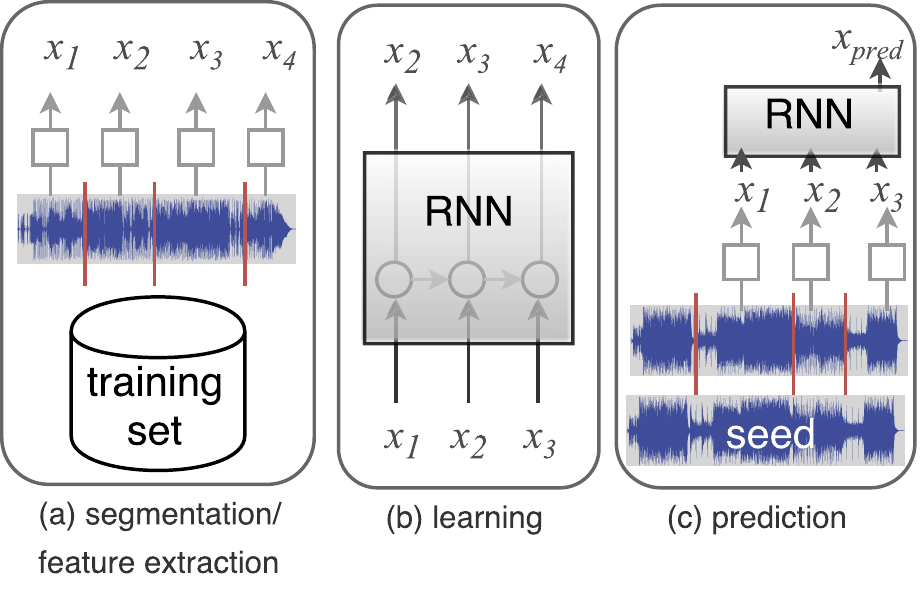}}
\caption{A block diagram of the proposed algorithm, (a,b) training of RNN and (c) prediction of a feature vector, $x_{pred}$.}
\label{figure:block}
\end{center}
\end{figure} 

%\subsection{Overview}
 Figure \ref{figure:block} illustrates the procedure of dataset construction, as well as the training and 
prediction stages of the proposed algorithm. 
First, the training tracks are segmented and $x_i$, the features 
for each segments are extracted (Fig. \ref{figure:block}a). Then an RNN of length $N$ ($N$=3 in the figure) is trained to learn the transitions of the sequence of feature vectors (Fig. \ref{figure:block}b). 
When a seed track is provided, the features of the last $N$ segments are extracted and fed into 
the trained RNN to predict the feature vector $x_{pred}$ (Fig. \ref{figure:block}c). 
The algorithm selects a track with a start segment that is most similar to $x_{pred}$.

\subsection{Structural Segmentation} Structural segmentation is a task
aiming to find the boundaries of different segments or parts in music, e.g.
\textit{intro, verse, bridge, chorus}. The most common approach
is to take advantage of self-similarity between frames of the track \cite{foote2000automatic}.
In the experiment, we used a basic and efficient method that is proposed in \cite{foote2000automatic}.
Although the results introduce some errors, the
feature vector sequences that are based on the imperfect segmentation
still approximate the information about
how each feature changes along time in each track.

\subsection{Feature Extraction} \label{sec:featext} The proposed algorithm can use 
feature extraction methods that are relevant to
listeners' musical preferences and able to represent a musical segment.
This includes estimated latent features from collaborative filtering \cite{van2013deep}, 
tags such as genre and emotion \cite{dieleman2014end}
or implicit features such as the weights of
the last hidden layer of a neural network classifier \cite{liangcontent}. 
Using explicit labels such as genre can facilitate explaining
the behaviour of the algorithm, which is important for research
and also to the listener. 
% We opt for this approach here. 

In the experiment, an auto tagging algorithm in \cite{choi2016automatic} is used to predict a 50-dimensional vector whose elements correspond to the probability of each tag. The tagging algorithm is based on deep convolutional neural networks and trained on Million Song Dataset \cite{bertin2011million}. It shows state-of-the-art performance while the tags cover variety of categories such as genre, emotion, instrument, and era.
Although some of the tags such as genre typically characterise the entire music track, they are not necessarily constant over the whole track. %(but usually they would be constant, as those labels are constant in general)

\subsection{RNN Model} The goal of RNN training is to predict
the feature of the following track ($x_{pred}$) that maintains consistency and fluctuations, i.e., a certain variation over the features.
To this end, a 2-layer RNN with 512 hidden units is employed. LSTM units 
 \cite{gers2000learning} are used as they show state-of-the-art performance among RNN variants for
 several sequence modelling tasks \cite{greff2015lstm}.

\subsection{Similarity Measure} 
A similarity measure is necessary to find the subsequent track using the feature vector predicted by the RNN. 
The similarity metric directly affects the properties 
of the generated playlists and therefore it should be carefully selected. 
Using the \textit{cosine distance} may compensate for the popularity bias and result
in recommending more niche items \cite{schedl2015music}. 
The \textit{Discounted Cumulative Gain} (DCG) turned out to be effective in our experiment.

DCG is a weighted version of \textit{Cumulative Gain} (CG). 
CG is designed to measure ranking quality of a retrieved list and  
DCG weights on the top-$N$ relevant items by \textit{discounting} lower relevant items.
Applying this measure was motivated by the type of feature extraction algorithm we use. 
The extracted feature is a vector of probabilities of each tag
and tags with large probabilities should be weighted more than the others 
to facilitate maintaining the consistency of the generated playlists.
Because DCG weights high-ranking elements more, it can theoretically work better.

%===============
% 3. RESULT/DISCUSSION
%===============

\section{Results and Discussion}
\label{sec:exp}
\subsection{Configurations}
For training, we used a private dataset with 28,430 commercial tracks of modern popular music including Rock, 
Hip-Hop, and Jazz.
We used 7,880 tracks from the \textit{ILM10k} dataset for testing \cite{saari2015genre} \cite{allik:16}. 

The segmentation is performed using \cite{foote2000automatic}, which is implemented in \cite{nietomsaf}.
As mentioned in Section \ref{sec:prop}, an
automatic tagging algorithm was used as a feature extractor \cite{choi2016automatic}. 
An RNN with a length of 50 is trained. We compared DCG with the cosine distance
and the $l_2$ norm for computing similarity. Audio processing and RNN are implemented using \textit{librosa} \cite{mcfee2015librosa}, \textit{Keras} \cite{chollet2015}, and \textit{Theano} \cite{team2016theano}.

\begin{figure}[t!]
\begin{center}
\centerline{\includegraphics[width=\columnwidth]{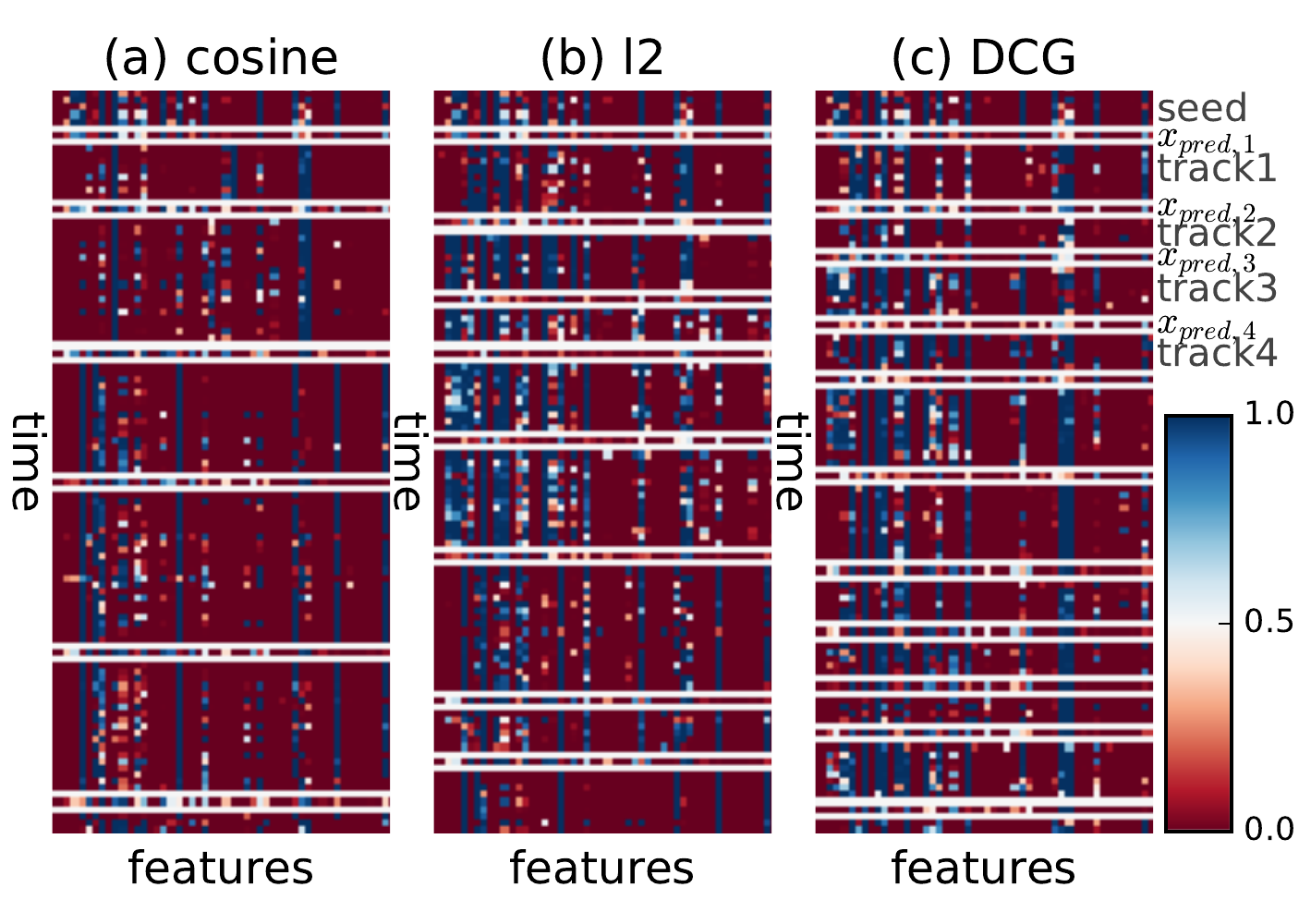}}
\caption{Transitions of feature vectors by cosine distance, $l_2$, and 
DCG on left, centre, and right, respectively. Y-axis is time (top to bottom)
and x-axis refers to the 50 feature dimensions. The first track (topmost 
feature vectors) is the seed.}
\label{fig:transitions}
\end{center}
\end{figure} 

Figure \ref{fig:transitions} shows the transitions of feature vectors
in three playlists given the same seed track but different similarity metrics. 
The seed track is represented by a 7-by-50 matrix, i.e., 7 segments of the track.
White horizontal lines indicate beginnings and ends of each track. 
The predicted feature vectors (1-by-50) are illustrated in between. 

The figure helps to explain several aspects of qualitative observations 
by the authors while listening to the generated playlists. 

\subsection{Discussions}
\textbf{First}, we found both consistency and fluctuations in 
the extracted features within tracks.
In general, several features show consistently large (blue) and small (red) values,
while the other features vary. It supports the selection of feature extraction algorithm. % NEW

However, there are still rooms for further improvements. For example, whitening of each feature dimension can be adopted to compensate the prior distributions of each feature. Although RNNs are able to adapt to such differences, the similarity measure may be affected from such pre-processing. % NEW

\textbf{Second}, the transitions usually successfully keep the coherence within playlists as demonstrated by the figure.
However, we noticed that the model is prone to missing overall similarity in long playlists.
It may be related to the observation that the trained RNN occasionally predicts a vector that does not have near neighbours. It means the selected track is not similar enough to the predicted vector, which may result in an undesirable or suboptimal transition. % NEW

This may first be due to the short lengths of segments in the training data. 
The majority of training tracks have fewer segments (90$\%$ of
the training tracks has less than 17 segments), therefore
the long-term dependency may not be learnt and the prediction
may be dominated by short-term features. 

In future work, training with longer sequences 
such as concatenated features of tracks from sequences in
an albums, setlists or curated playlists may be used to help learning better transitions.
It can be also a typical behaviour of RNNs. Although RNNs generally model
the long-term dependency of sequences, in many cases, RNNs have shown a 
behaviour puts more emphasis on recent inputs rather than older ones. 

However, the problem seems to be partly resolved when DCG is used as the similarity measure as discussed alongside out last set of observations.

\textbf{Third}, DCG provides more coherent playlists compared to the cosine distance and $l_2$-norm.
This phenomenon is found not only in the sequences corresponding to Figure \ref{fig:transitions} but also in other playlists. This can be explained as follows. In each track, there are consistently strong, consistently weak, and fluctuating features. This pattern, especially the consistencies, can be easily learned by the RNN and the consistent features are maintained in the predictions. Finally, DCG prioritises the features that are large in predictions, resulting in successfully finding a track with those consistently large features. This improves the coherence of the resulting playlists.

%Although such a property, that weights certain
%dimensions, can be heuristically implemented for other similarity measures, 
%DCG may be more flexible   % TODO

\section{Conclusion}
\label{sec:conclusinon}

In this paper, we proposed a novel algorithm for playlist generation that relies on learning desirable musical qualities from within-track transitions between musical segments. The proposed combination of an RNN, within-track structure and DCG showed an encouraging result. Within-track structures showed the consistency and dynamics that are assumed in playlists. An RNN model learnt the feature sequences and its predictions are successfully used for the selections of following tracks. Different similarity measures resulted in different playlists. 

Future work will investigate advanced architectures such as bidirectional RNNs \cite{schuster1997bidirectional} and more formal assessments. 
Using bidirectional RNNs can be used to create playlists that have more constraints e.g. start and end tracks \cite{flexer2008playlist} and steerability \cite{maillet2009steerable}. Formal assessments will include subjective measurement e.g. satisfaction and objective measures with regards to consistency, fluctuations and diversity. % NEW

\bibliographystyle{acm} 
\bibliography{icml_2016_workshop_playlist}

\end{document}